\pdfoutput=1

\documentclass[11pt]{article}

\usepackage{naacl2021}

\usepackage{times}
\usepackage{latexsym}

\usepackage[T1]{fontenc}

\usepackage[utf8]{inputenc}

\usepackage{microtype}

\usepackage{hyperref}       
\usepackage{url}            
\usepackage{booktabs}       
\usepackage{amsfonts}       
\usepackage{nicefrac}       
\usepackage{microtype}      

\usepackage{graphicx}
\usepackage[font=small]{caption}
\usepackage{subcaption}
\usepackage{wrapfig}

\usepackage{amsfonts}
\usepackage{amsmath}
\usepackage{amssymb}
\usepackage{amsthm}

\usepackage{algorithmic}
\usepackage{algorithm}

\usepackage{booktabs}
\usepackage{multirow}
\usepackage{tabularx}

\usepackage{xspace}
\usepackage{enumitem}

\usepackage{scrextend}

\usepackage{macro}

\newcommand*{\escape}[1]{\texttt{\textbackslash#1}}

\definecolor{stage-orange}{RGB}{242, 114, 6} 
\definecolor{stage-green}{RGB}{1, 113, 0} 

\newcommand\PROG{ProGen\xspace}
\newcommand\GPT{GPT-2\xspace}
\newcommand\BART{BART\xspace}

\newcolumntype{I}{!{\vrule width 1pt}}

\title{Progressive Generation of Long Text \\ with Pretrained Language Models}

\author{
Bowen Tan$^1$,~~
Zichao Yang$^1$,~~
Maruan Al-Shedivat$^1$,~~
Eric P. Xing$^{1,2,3}$,~~
Zhiting Hu$^{1,4}$\\
$^1$Carnegie Mellon University,~~ $^2$Petuum Inc.,~~ $^3$MBZUAI,~~ $^4$UC San Diego\\
{\small 
{\tt \{btan2, zichaoy, alshedivat, epxing\}@andrew.cmu.edu, zhh019@ucsd.edu}
}
}

\begin{document}
\maketitle

\begin{abstract}
Large-scale language models (LMs) pretrained on massive corpora of text, such as GPT-2, are powerful open-domain text generators.
However, as our systematic examination reveals, it is still challenging for such models to generate coherent long passages of text (e.g., 1000 tokens), especially when the models are fine-tuned to the target domain on a small corpus. Previous planning-then-generation methods also fall short of producing such long text in various domains.
To overcome the limitations, we propose a simple but effective method of generating text in a progressive manner, inspired by generating images from low to high resolution.
Our method first produces domain-specific content keywords and then progressively refines them into complete passages in multiple stages.
The simple design allows our approach to take advantage of pretrained LMs at each stage and effectively adapt to any target domain given only a small set of examples.
We conduct a comprehensive empirical study with a broad set of evaluation metrics, and show that our approach significantly improves upon the fine-tuned large LMs and various planning-then-generation methods in terms of quality and sample efficiency. Human evaluation also validates that our model generations are more coherent.\footnote{{Code available at \url{https://github.com/tanyuqian/progressive-generation}}}
\end{abstract}

\section{Introduction}
\label{sec:introduction}

\begin{figure}[!t]
    \centering
    \includegraphics[width=0.5\textwidth]{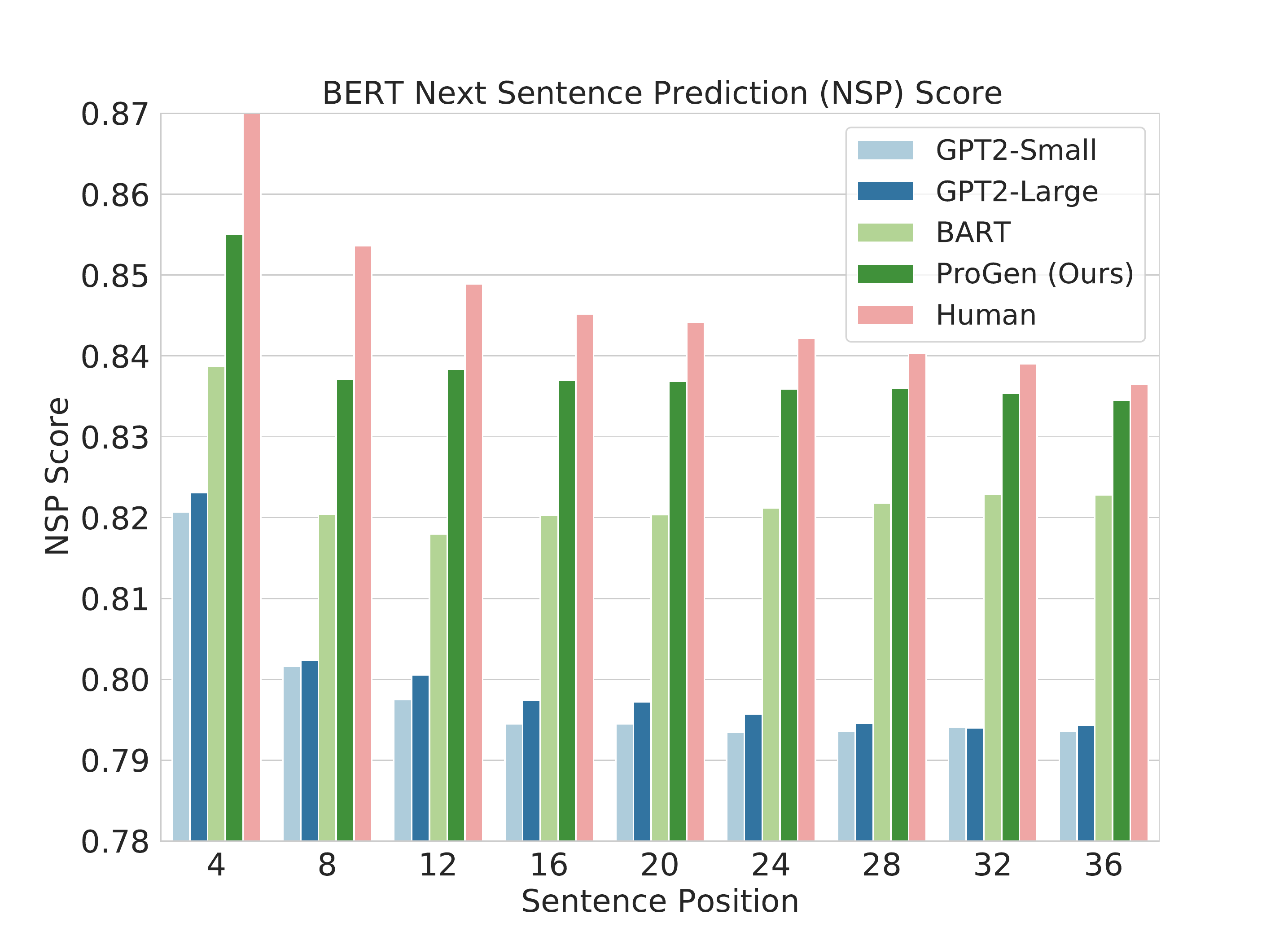}
    \vspace{-20pt}
    \caption{
    Results of large-scale LMs (\GPT and BART) fine-tuned on 10K stories.
    Coherence of text is evaluated by BERT next sentence prediction (NSP) score, where x-axis is the position of the evaluated sentences in the passage.
    There is a significant gap
    in coherence between text by human and text by large-scale LMs. Our proposed \PROG
    instead generates more coherent samples close to human text.
    }
    \label{fig:bert_nsp_story}
\end{figure}

Generating coherent long text (e.g., 1000s of tokens) is useful in myriad applications of creating reports, essays, and other long-form content. Yet the problem is particularly challenging as it demands models to capture global context, plan content, and produce local words in a consistent manner. Prior studies on ``long'' text generation have typically limited to outputs of 50-200 tokens~\citep{shen2019towards,bosselut2018discourse,zhao2020graph}.

Recent large-scale pretrained language models (LMs), such as GPT-2~\citep{radford2019language} and BART~\citep{lewis2019bart}, emerged as 
an impressive open-ended text generator capable of producing surprisingly fluent text. The massive LMs are typically pretrained on large corpora of generic text once, and then fine-tuned with small domain-specific data.
%
%
The latest work has mostly focused on the regime of relatively short text with low hundreds of tokens. 
For example, \citet{holtzman2019curious,see2019massively,hua2020pair} studied \GPT and \BART generations with a maximum length ranging from 150 to 350 tokens. In this work, we study the problem of generating coherent, much longer passages of text (e.g., 1000 tokens).
GPT-3~\citep{brown2020language} was reported to produce long essays, yet the results seem to need extensive human curations~\citep[e.g.,][]{gpt3exposed,gpt3guardian}, and the system is not publicly available to adapt to arbitrary desired domains.

In this work, we examine fine-tuning of large-scale LMs for domain-specific generation of 
extra-long text.
We find that samples produced by \GPT fine-tuned on small domain-specific corpora exhibit various imperfections, including excessive repetitiveness and incoherence between sentences far apart. Figure~\ref{fig:bert_nsp_story} measures the coherence of text generated by the fine-tuned \GPT w.r.t the BERT next sentence prediction~\cite{devlin2018bert} score.
As the figure shows, \GPT models (regardless of the model size) exhibit a significant gap in the score compared with human text, hence falling short in generating coherent text. 

We hypothesize that the problem is mainly caused by the sequential generation order of the LMs, which makes global content planning of the passage difficult, especially when the generated text is long and contains thousands of words. One could potentially adopt the recent \emph{planning-then-generation} or \emph{non-monotonic} methods (Sec~\ref{sec:discussion-related-work}), yet those methods either require specialized neural architectures that need costly retraining for each domain~\cite{gu2019insertion,stern2019insertion,chan2019kermit,fan2019strategies}, or rely on dedicated intermediate content plans (e.g., summaries, SRL labels)~\cite{fan2019strategies,yao2019plan} with limited flexibility and producing sub-optimal results as shown in our experiments.

\begin{figure*}[t]
    \centering
    \includegraphics[width=\textwidth]{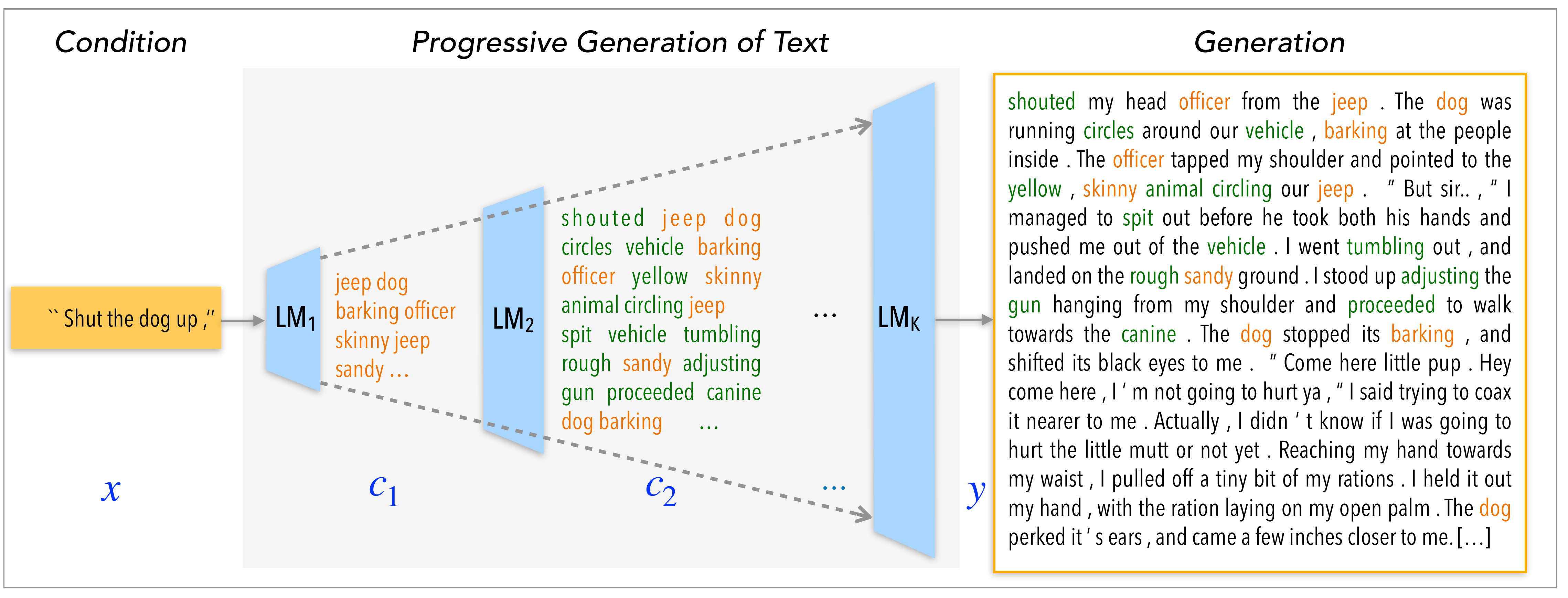}
    \caption{
    Progressive generation of long text $\yv$ given any condition $\xv$. Each stage refines the results from the previous stage by adding finer-grained details. Added content at each stage is highlighted in different colors.
    }
    \label{fig:architecture}
\end{figure*}

To overcome the limitations, 
we introduce a new method for \textbf{Pro}gressive \textbf{Gen}eration of Text (\PROG). 
We observe that generation of some words (e.g., stop words) does not require many contexts, while other words are decisive and have long-term impact on the whole content of the passage.
Motivated by this observation, our approach first produces a sequence of most informative words, then progressively refines the sequence by adding finer-grained details in multiple stages, until completing a full passage. The generation at each stage is conditioning on the output of the preceding stage which provides anchors and steers the current generation (Figure~\ref{fig:architecture}). The intermediate words produced at each stage are defined based on a simple TF-IDF informativeness metric. 

The approach enjoys several core advantages: {\bf (1)} Although the progressive approach implements a conceptually non-monotonic generation process, generation at each stage can still be performed in a left-to-right manner and thus is directly compatible with the powerful pretrained monotonic LMs. The LMs at different stages are easily fine-tuned to accommodate a target domain using only small, independently constructed data. Intuitively, each LM is addressing a sub-task of mapping a sequence to a finer-resolution one,
which is much simpler than the overall task of mapping from conditions to full passages of text. 
In this work, we use \BART~\citep{lewis2019bart} for generation at each stage, though one can also plug in other off-the-shelf LMs. As seen from Figure~\ref{fig:bert_nsp_story},
\PROG can generate more much coherent text compared with \GPT and nearly match human text in terms of the BERT-NSP score; {\bf (2)} In contrast to the typical 2-stage planning-then-generation in prior work, the simple progressive strategy offers added flexibility for an arbitrary number of intermediate stages, yielding improved results; {\bf (3)} The training data for each stage is extracted from domain corpus using the simple TF-IDF metric, without need of additional resources (e.g., pretrained summarization models) as in prior work, making the method broadly applicable to various domains and languages.

We conduct extensive empirical studies on the CNN News \citep{hermann2015teaching} and WritingPrompts \citep{fan2018hierarchical} corpora, evaluating various systems by a wide-range of automatic metrics as well as human judgement. Results show that \PROG achieves strongly improved performance by decomposing the generation into more progressive stages. Our method produces diverse text passages of higher quality and coherence than a broad set of models, including fine-tuned \GPT, \BART, and other various planning-then-generation strategies.

\section{Related Work}
\label{sec:discussion-related-work}

\paragraph{Content planning in generation.}
The idea of separate content planning and surface realization has been studied in early text generation systems~\citep{reiter1997building}. 
Recent neural approaches have also adopted similar planning-then-generation strategies for data-to-text~\cite{moryossef2019step,puduppully2019data}, storytelling~\cite{fan2019strategies,yao2019plan,xu2020megatron}, machine translation~\cite{ford2018importance}, and others~\cite{hua2019sentence,yao2017towards}. These models often involve customized architectures incompatible with the existing large LMs. Scaling those models for long text generation thus can require expensive training, which restricts systematic studies. On the other hand, it is possible to adopt some of the content planning strategies (e.g., summaries or SRL sequences as the plans~\citep{fan2019strategies}), and repurpose pretrained LMs for generation in each stage. However, these strategies with dedicated intermediate plans and a pre-fixed number (typically 2) of stages can have limited flexibility, leading to sub-optimal results as shown in our empirical study. Besides, creating training data for planning requires additional resources (e.g., pretrained summarization models or SRL models) which are not always available (e.g., in certain domains or for low-resource languages).
In contrast, we propose a simple way for designing the intermediate stages based on word informativeness, which can flexibly increase the number of stages for improved results, and easily create training data for all stages without additional models.

\paragraph{Non-monotonic generation and refinement.} 
Another relevant line of research is non-monotonic generation~\cite{welleck2019non,gu2019insertion,stern2019insertion,chan2019kermit,zhang2020pointer}, infilling~\cite{zhu2019text,shen2020blank,qin-etal-2020-back}, or refinement~\cite{lee2018deterministic,novak2016iterative,mansimov2019generalized,kasai2020disCo} that differs from the restricted left-to-right generation in conventional LMs. Again, those approaches largely depend on specialized architectures and inference, making them difficult to be integrated with the powerful pretrained LMs. The prior studies have focused on generating short text. 
Our proposed coarse-to-fine progressive generation conceptually presents a non-monotonic process built upon the pretrained monotonic LMs, which permits fast adaptation to any target domain and generation of much longer text.


\paragraph{Long text generation.}
Previous work has made attempts to generate text of up to two or three hundred tokens. Those methods often adopt the similar idea of planning-then-generation as above~\citep{shen2019towards,zhao2020graph,bosselut2018discourse,see2019massively,hua2020pair,rashkin-etal-2020-plotmachines}. Another line of work instead focuses on extending the transformer architecture~\cite{vaswani2017attention} to model longer text sequences~\cite[e.g.,][etc]{dai2019transformer,wang2020linformer,choromanski2020rethinking}. 
For example, \citet{liu2018generating} used a hybrid retrieval-generation architecture for producing long summaries; \citet{dai2019transformer} showed long text samples qualitatively. 
Our work systematically examines the pretrained LMs in generating long domain-specific text, and proposes a new approach that empowers pretrained LMs for producing samples of significantly higher-quality.

\section{Progressive Generation of Text}
\label{sec:methods}

One of the main challenges in generating long coherent passages is modeling long-range dependencies across the entire sequences (e.g., 1000 tokens). We propose a progressive generation approach that is conceptually simple yet effective. 
Intuitively, progressive generation divides the complex problem of generating the full passage into a series of much easier steps of generating coarser-grained intermediate sequences.
Contrary to generating everything from left to right from scratch, our progressive generation allows the model to first plan globally and then shift attention to increasingly finer details, which results in more coherent text. Figure~\ref{fig:architecture} illustrates the generation process.





\subsection{Generation Process}\label{sec:method:gen}

Let $\yv := [y_1, y_2, \dots, y_T]$ be the output text, where each $y_i$ is a token of language (a word or a sub-word). 
The output sequences are generated either conditionally on any other information $\xv$ (\eg, generations of a story given a prompt), or unconditionally (in which case we assume $\xv \equiv \varnothing$ while keeping the same notation).

Instead of generating the full passage $\yv$ directly, we propose to add multiple intermediate stages: 
$\xv \rightarrow \cv_1 \rightarrow \cv_2  \cdots \rightarrow \cv_{K} \rightarrow \yv$, 
where for each stage $k\in\{1,\dots,K\}$, $\cv_k$ is an intermediate sequence containing information of the passage at certain granularity. For instance, at the first stage, $\cv_1$ can be seen as a highest-level content plan consisting of the most informative tokens such as key entities. 
Then, based on the plan, we gradually refine them into subsequent $\cv_k$, each of which contains finer-grained information than that of the preceding stage. At the final stage, we refine $\cv_K$ into the full passage by adding the least informative words (e.g., stop words).
%
The generation process corresponds to a decomposition of the conditional probability as: 
\begin{align}
\small
     & \prob{\yv, \{\cv_k\} | \xv} = \prob{\cv_1|\xv} \nonumber\\
     & \quad\quad\quad \Pi_{k=2}^K \prob{\cv_k|\cv_{k-1}, \xv} \prob{\yv |\cv_{K}, \xv}.
\end{align}
As the above intuition, $\cv_k$ at early stages as the high-level content plans should contain informative or important words, to serve as skeletons for subsequent enrichment. 

We next concretely define the order of generation, namely, which words should each stage generates. Specifically, we propose a simple method that constructs a vocabulary $\mathcal{V}_k$ for each stage $k$, based on the \emph{importance} of words in the target domain. Each particular stage $k$ only produces tokens belonging to its vocabulary $\mathcal{V}_k$. By the progressive nature of the generation process, we have $\mathcal{V}_1\subset \cdots\subset \mathcal{V}_K\subset \mathcal{V}$. That is, $\mathcal{V}_1$ contains the smallest core set of words in the domain, and the vocabularies gradually expand at later stages until arriving the full vocabulary $\mathcal{V}$. 
Note that vocabularies in later stages are supersets of those in earlier stages. This allows the later stages to remedy and polish potential mistakes made in earlier stages when necessary.
We discuss the construction of the vocabularies in the below.


\paragraph{Stage-wise vocabularies based on word importance.} \label{sec:stage-vocab}
Given a text corpus $\mathcal{D}$ of the target domain with the full vocabulary $\mathcal{V}$, we define the importance scores of words in $\mathcal{V}$ based on the TF-IDF metric. We then rank all the words and assign the top $V_k$ words to the intermediate vocabulary $\mathcal{V}_k$. Here $V_k$ is a hyper-parameter controlling the size of $\mathcal{V}_k$. 

More concretely, for each word $w\in\mathcal{V}$, we first compute its standard TF-IDF score \cite{salton1986introduction} in each document $\dv\in\mathcal{D}$, which essentially measures how important $w$ is to $\dv$.
The importance of the word $w$ in the domain is then defined as the average TF-IDF score across all documents containing $w$:
\begin{equation}
\small
    \label{eq:term-importance}
    \mathrm{importance}(w, \mathcal{D}) = \frac{\sum_{\dv \in \mathcal{D}} \mathrm{TF\_IDF}(w, \dv)}{\mathrm{DF}(w, \mathcal{D})},
\end{equation}
where $\mathrm{TF\_IDF}(w, \dv)$ is the TF-IDF score of word $w$ in document $\dv$; and $\mathrm{DF}(w, \mathcal{D})$ is the document frequency, \ie, the number of documents in the corpus that contain the word $w$.

\begin{algorithm}[t]
\small
	\caption{\small Training for Progressive Text Generation}
	\label{alg:progressive-training}
 	{\bf Inputs:} \\
 	Domain corpus $\mathcal{D}$ \\
 	Vocabulary sizes for $K$ stages \\
    $K$ pretrained LMs (\eg~\GPT or BART)\\
	\begin{algorithmic}[1]
        \STATE Construct stage-wise vocabularies $\{\mathcal{V}_k\}$ based on word importance Eq.\eqref{eq:term-importance}
        \STATE Extract intermediate sequences $\{\cv^*_k\}$ using $\{\mathcal{V}_k\}$; add data noises (Sec~\ref{sec:method:train})
        \STATE Fine-tune all LMs independently (Sec~\ref{sec:method:train}) \\

	\end{algorithmic}
	\ \\
	 {\bf Output:} Fine-tuned LMs for generation at all stages in a progressive manner
\end{algorithm}


\paragraph{Pretrained language models as building blocks.}
Compared to many of the previous planning-then-generation and non-monotonic generation methods, one of the key advantages of our progressive generation design is the direct compatibility with the powerful pretrained LMs that perform left-to-right generation. Specifically, although our approach implements a non-monotonic generation process that produces importance words first, we can generate intermediate sequences $\cv_k$ at each stage still in a left-to-right manner. Thus, we can plug pretrained LM, such as \GPT or BART, into each stage to carry out the generation. As described more in section~\ref{sec:method:train}, for each stage $k$, we can conveniently construct stage-specific training data from the domain corpus $\mathcal{D}$ using the stage-wise vocabulary $\mathcal{V}_k$, and fine-tune the stage-$k$ LM in order to generate intermediate sequences at the stage that are pertaining to the target domain. 

One can add masks on the pretrained LM's token distributions to ensure the stage-$k$ LM only produces tokens belonging to $\mathcal{V}_k$. In practice, we found it is not necessary, as the pretrained LM can usually quickly learns the pattern through fine-tuning and generate appropriate tokens during inference. In our experiments we use BART for all stages, since BART is an encoder-decoder model which can conveniently take as inputs the resulting sequence from the preceding stage and generate new. (For the first stage in an unconditional generation task, we simply set $\xv=\varnothing$.)
We note that \GPT, and other relevant pretraiened LMs, can indeed also be used as a conditional generator~\cite{radford2019language,liu2018generating} and thus be plugged into any of stages.

\subsection{Training}\label{sec:method:train}
Our approach permits straightforward training/fine-tuning of the (pretrained) LMs at different stages given the domain corpus $\mathcal{D}$. In particular, we can easily construct independent training data for each stage, and train all LMs in parallel. 
Note that no additional resources such as pretrained summarization or semantic role labeling models are requested as in previous work, making our approach directly applicable to a potentially broader set of domains and languages. We plan to explore the use of our method in multi-lingual setting in the future.

More concretely, for each stage $k$, we use the stage vocabularies $\mathcal{V}_{k-1}$ and $\mathcal{V}_k$ to filter all relevant tokens in the documents as training data. That is, given a document, we extract the sub-sequence $\cv_{k-1}^*$ of all tokens from the document that are belonging to $\mathcal{V}_{k-1}$, and similarly extract sub-sequence $\cv_{k}^*$ belonging to $\mathcal{V}_{k}$. The $\cv_{k-1}^*$ and $\cv_{k}^*$ are then used as the input and the ground-truth output, respectively, for training the LM at stage $k$ with maximum likelihood learning. Therefore, given the stage-wise vocabularies $\{\mathcal{V}_k\}$, we can automatically extract training data from the domain corpus $\mathcal{D}$ for different stages, and train the LMs separately. 

In the multi-stage generation, the intermediate sequences are not natural language. Yet we found that fine-tuning pretrained LMs (such as BART and GPT-2) to generate the intermediate sequences is indeed very efficient in terms of data and computation. We tried training other models such as small sequence-to-sequence models and n-gram models from scratch, which we found is much harder, requiring more data, or yielding inferior performance. This again highlights the importance of using pretrained LMs, as enabled by our simple method design.

\textbf{Stage-level exposure bias and data noising.}
In the above training process, the outputs of each LM are conditioning on the ground-truth input sequences extracted from the real corpus. In contrast, at generation time, the LM takes as inputs the imperfect sequences produced at the previous stage, which can result in new mistakes in the outputs since the LM has never be exposed to noisy inputs during training.
Thus, the discrepancy between training and generation can lead to mistakes in generation accumulating through the stages. The phenomenon resembles the \emph{exposure bias} issue~\cite{ranzato2015sequence} of sequential generation models at token level, where the model is trained to predict the next token given the previous ground-truth tokens, while at generation time tokens generated by the model itself are instead used to make the next prediction.

To alleviate the issue and increase the robustness of each intermediate LM, we draw on the rich literature of addressing token-level exposure bias~\cite{xie2017data,tan2018connecting}. Specifically, during training, we inject noise into the ground-truth inputs at each stage by randomly picking an $n$-gram ($n\in\{1,2,3,4\}$) and replacing it with another randomly sampled $n$-gram. The data noising encourages the LMs to learn to recover from the mistakes in inputs, leading to a more robust system during generation. 

\section{Experiments}
\label{sec:experiments}


\subsection{Setup}

\paragraph{Domains.} 
We evaluate on two text generation domains including: {\bf (1) CNN News} \cite{hermann2015teaching} for unconditional generation. 
{\bf (2) WritingPrompts} \cite{fan2018hierarchical} for conditional story generation. The task is to generate a story given a prompt. 
The two datasets are chosen since they both contain long documents, with 
CNN's average and maximum length being 512 and 926, and WritingPrompts's 
being 437 and 942, respectively.
To demonstrate the data efficiency of our approaches adapting to target domains,
we sample 1,000 documents in each dataset for training.

\paragraph{Model configs.} 
We use BARTs for all stages of generation. 
Due to computation limitations, we experiment models with 2, 3, 4-stages
generations. In our 2-stage model, our first stage covers about 25\% of all content; 
in the 3-stage model, the first and second stages cover 15\% and 25\% of all content, respectively; and in the 4-stage model, our first three stages cover 15\%, 20\%, 25\%
of all content. 
For model training, we follow the same protocol as \cite{see2019massively} to fine-tune all pretrained models until convergence. To combat exposure bias, we add noise to the training data as described in Sec~\ref{sec:method:train}, with the probability of replacing 1,2,3,4-grams 0.1/0.05/0.025/0.0125.
In the generation phase, we use top-p decoding \cite{holtzman2019curious} with $p=0.95$ to generate 1024 tokens at maximum. 
Experiments were conducted with RTX6000 GPUs. It took around 4 hours for model fine-tuning and generation with a single GPU.

\paragraph{Comparison methods.}
We compare with a wide range of baselines, categorized into two groups: {\bf (1)} The large {pretrained LMs} including BART~\cite{lewis2019bart} and GPT-2 in both small and large sizes~\citep{radford2019language}. The LMs generate text in a standard left-to-right manner; {\bf (2)} Progressive generation with various strategies adopted in the prior planning-then-generation work. Same as our proposed method, each stage adapts a pretrained BART for generation. Specifically, {\bf Summary} first generates a short summary text as the content plan and conditioning on the summary produces the full passage of text~\cite{fan2019strategies}. For training, summaries are obtained using the state-of-the-art pretrained CNN news summarization model based on BART; {\bf Keyword} first generates a series of keywords, based on which the full text is generated in the next stage. Following~\citep{yao2019plan}, the keywords are extracted with the RAKE algorithm~\cite{rose2010automatic} for training; {\bf SRL} follows the recent work~\citep{fan2019strategies} by first generating a sequence of predicates and arguments and then producing the full text conditionally. The same semantic role labeling tool as in the prior work
is used here to create training data. {\bf SRL+NER} and {\bf SRL+Coref} further augment the SRL method by an additional stage of generating entity anonymized text conditioning on the predicates sequence prior to the final stage \citep{fan2019strategies}. SRL+NER uses an NER model to mask all entities, while SRL+Coref applies coreference resolution to mask all clusters of mentions. We use the same NER and coreference tools as in \citep{fan2019strategies}. Finally, as a reference, we also present the results of {\bf Human}-written text (i.e., the text in the dev set).

\subsection{Automatic Evaluation}

\subsubsection{Evaluation Metrics}

To evaluate the generation quality for the domain-specific open-ended generation as studied here, we primarily measure the ``closeness'' between two \textbf{sets} of text, one generated by the model and the other the real text from the target domain.
We evaluate with a broad array of automatic metrics, including \emph{lexical-based quality} metrics and \emph{semantic-based quality} metrics. We also evaluate the generation \emph{diversity}. 

\paragraph{MS-Jaccard (MSJ)} is a lexical-based metric \cite{montahaei2019jointly}, where MSJ-$n$ measures the similarity of $n$-grams frequencies between two sets of text with Jaccard index. 

\vspace{-4pt}

\paragraph{TF-IDF Distance (TID)} is defined as the distance between the average TF-IDF features of two text sets. We use it as an additional lexical-based quality measure.

\vspace{-4pt}

\paragraph{Fr\'echet BERT Distance (FBD)} is a semantic-based metric \cite{montahaei2019jointly} that measures the Fr\'echet Distance in the BERT feature space between the generated and real text. By using the BERT features from shallow (S), medium (M), and deep (D) layers, we can compute FBD-S/M/D, respectively.

\vspace{-4pt}

\paragraph{Backward BLEU (B-BLEU)} is a diversity metric \cite{shi2018toward} measuring how well the generated text covers n-grams occurred in the test set.

\vspace{-4pt}

\paragraph{Harmonic BLEU (HA-BLEU)} \cite{shi2018toward} is an aggregated quality and diversity metric that incorporates both the standard BLEU (i.e., precision) and the Backward BLEU (i.e., recall). 

\begin{figure*}[t]
    \centering
    \includegraphics[width=1.05\textwidth]{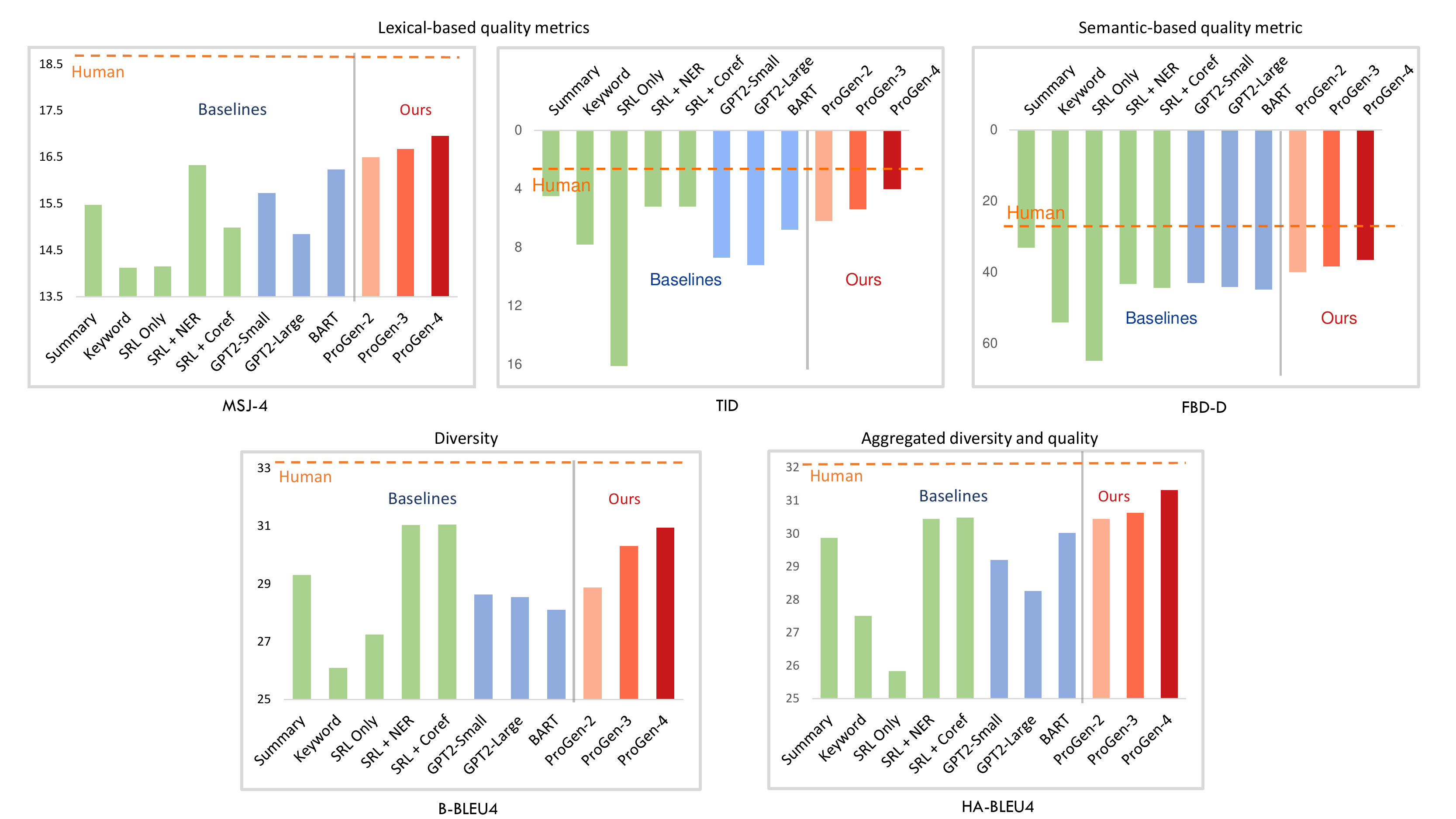}
    \vspace{-16pt}
    \caption{
    Results on the CNN News domain measured by different metrics. 
    For TID and FBD, the lower value the better.
    More results (MSJ-$n$, B-BLEU$n$ and HA-BLEU$n$ with different $n$ values, and FBD-S/M) are included in the appendix. 
    The three sets of comparison methods are shown in different colors, with our \texttt{\PROG} in {\color{red} red}, standard large LMs in {\color{blue} blue}, and the various models with previous planning strategies in {\color{nice-green} green}. Human results are shown as dashed lines, often indicating the best potential performance (except for the diversity related metrics).}
    \label{fig:cnn-results}
\end{figure*}
\begin{figure*}[!h]
    \centering
    \includegraphics[width=\textwidth]{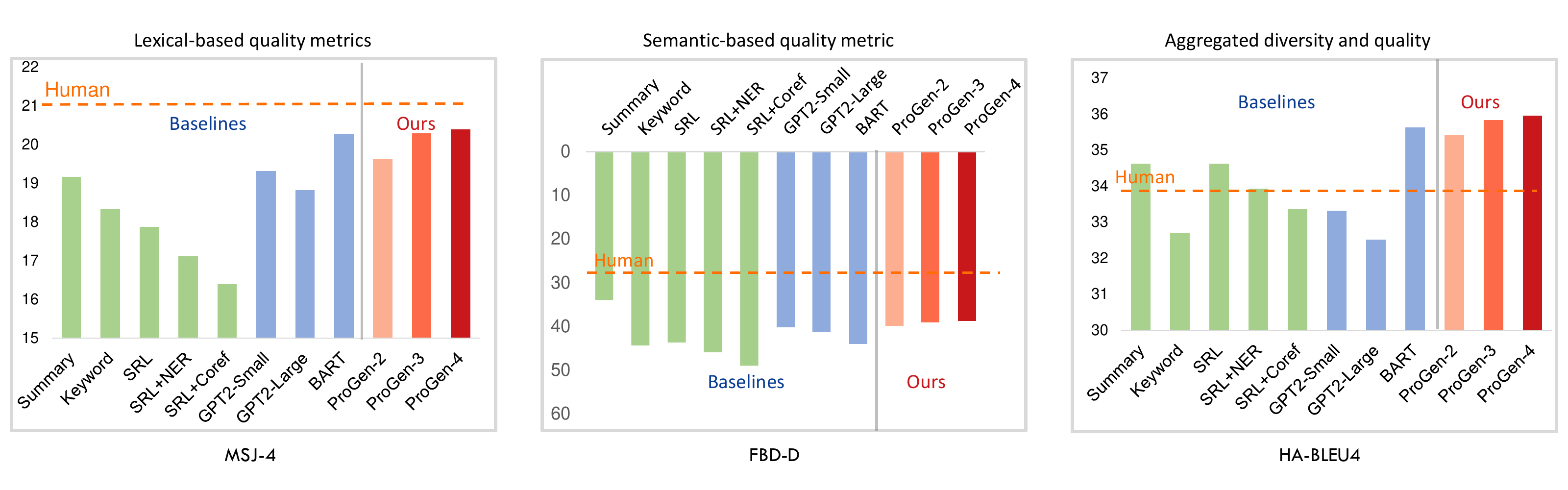}
    \caption{Results on the story domain measured by different metrics. More complete results are in appendix.}
    \label{fig:story-results}
\end{figure*}

\subsubsection{Results}
Figures~\ref{fig:cnn-results} and \ref{fig:story-results} show the results of the various systems on the news and story domains, respectively, measured with different metrics against test set. We give more complete results in the appendix. 
We can see that our progressive generation approach consistently outperforms the standard, single-stage LMs (\texttt{GPT2-Small}, \texttt{GPT2-Large} and \texttt{BART}) by a large margin on almost all metrics in both domains. Further, by increasing the number of progression stages, our method steadily achieves even stronger performance. This highlights the benefits of the flexible progressive generation strategy. 

The various models using pretrained LMs with previous planning-then-generation strategies show \emph{mixed} results across the different metrics. For example, \texttt{Summary} achieves strong performance in terms of the semantic-based quality metric FBD-D (partially because the summaries are closer to the real text in the BERT feature space), but significantly falls behind other models in terms of diversity (B-BLEU4) and other quality metrics like MSJ and HA-BLEU. 
Similarly, the \texttt{SRL}-based methods give only mediocre results in terms of the semantic-based FBD-D. In contrast, our approach maintains a relatively consistent performance level. In particular, our 4-stage model, \texttt{\PROG-4}, is steadily among the best across all metrics, further validating the advantage of the proposed simple yet flexible multi-stage generation.

These results also indicate the necessity of using a large diverse set of automatic metrics for a comprehensive evaluation, and motivate human studies for further assessment.

\begin{table}
\small
    \centering
    \begin{tabular}{@{}rlll@{}}
\toprule
           & \multirow{2}{*}{Fluency} & \multicolumn{2}{c}{Coherence} \\
           &  & passage & sentence (\%) \\ \midrule
BART       &    \textbf{4.59}     & 2.95                    & 70.29                    \\
GPT2-Small &    4.42     & 3.41                    & 74.69                    \\
Summary    &    4.39     & 3.37                    & 76.19                    \\ 
\PROG-4 (Ours)  &    4.46     & \textbf{3.83}           & \textbf{86.22}           \\ \bottomrule
\end{tabular}
\vspace{-6pt}
    \caption{Human evaluation results on CNN.}
    \label{tab:human_eval}
\vspace{-8pt}
\end{table}

\addtolength{\tabcolsep}{-1pt}    
\begin{figure*}
    \centering
    %
    %
    \begin{minipage}{0.28\textwidth}
    \vspace{-0.2cm}
        \centering
        \includegraphics[width=\textwidth]{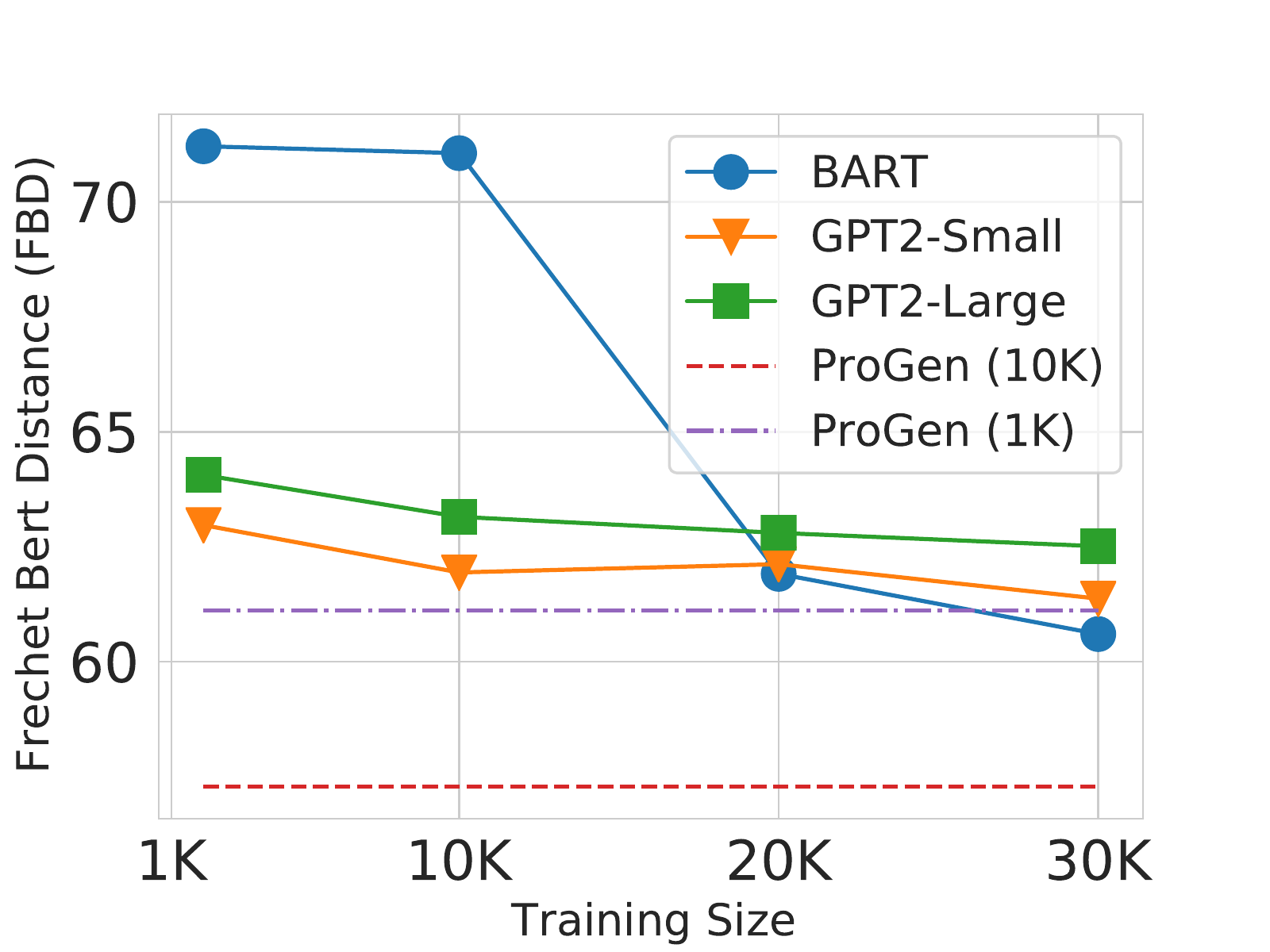}
        \vspace{-15pt}
        \captionof{figure}{Sample efficiency on the story domain with the FBD metric (the lower, the better).}
    \label{fig:sample_efficiency}
    \end{minipage}
    \hfill
    \begin{minipage}{0.35\textwidth}
        \small
        \centering
        \begin{tabular}{@{} r r r r r r @{}}
            \toprule
                     &  FBD-D $\downarrow$ & MSJ-4 $\uparrow$  & HA-BL4 $\uparrow$ \\ \midrule
            \PROG-2 &  \textbf{39.94} & \textbf{16.50}  & 30.45 \\
            -Noise   &  47.18 & 16.25  & \textbf{31.39} \\ \midrule
            \PROG-3 &  \textbf{38.30} & \textbf{16.68}   & 30.64 \\
            -Noise   &  39.64 & 16.65  & \textbf{30.72} \\ \midrule
            \PROG-4 &  \textbf{36.49} & \textbf{16.96}  & \textbf{31.32} \\
            -Noise   &  39.78 & 16.85   & 30.86 \\ \midrule
        \end{tabular}
    \captionof{table}{Effect of noise on CNN.}
        \vspace{1pt}
        \label{tab:noise}
    \end{minipage}
    \hfill
    \begin{minipage}{0.3\textwidth}
        \centering
        \small
        \begin{tabular}{@{}rcc@{}}
        \toprule
                        & FBD-D $\downarrow$ & TID $\downarrow$   \\ \midrule
        \PROG-2        & 39.94 & 6.2   \\
        GoldPlan        & 30.16 & 3.5   \\
        Human           & 25.63 & 2.6  \\ \bottomrule
        \end{tabular}
        \vspace{5pt}
            \captionof{table}{GoldPlan Results on CNN.}
            \label{tab:gold_plan}
    \end{minipage}
\end{figure*}

\begin{table*}[th!]
    \centering
    \small
    \begin{tabular}{p{\textwidth}@{}l@{}}
\toprule
 \textbf{First Stage:} {\color{stage-orange} beckham} \escape{n} {\color{stage-orange} liverpool bayern chelsea} \escape{n} {\color{stage-orange} beckham chelsea mancini} \escape{n} {\color{stage-orange} liverpool} \escape{n} {\color{stage-orange} club} [...] \\
 \textbf{Second Stage:} {\color{stage-orange} beckham} {\color{stage-green} newcastle} \escape{n} {\color{stage-green} footballer acquisition transfer} {\color{stage-orange} liverpool bayern} {\color{stage-green} munich} {\color{stage-orange} chelsea} {\color{stage-green} munich} \escape{n} {\color{stage-orange} beckham chelsea} {\color{stage-green} season italian siro gianluigi} {\color{stage-orange} mancini} {\color{stage-green} season italy} \escape{n} {\color{stage-green} season} {\color{stage-orange} liverpool} {\color{stage-green} season} \escape{n} {\color{stage-green} wembley} {\color{stage-orange} club} \escape{n} [...]
 \\
 \textbf{Final Stage:}\\
 (CNN) -- England striker Ashley {\color{stage-orange} Beckham} has joined Premier League strugglers {\color{nice-green} Newcastle} United. \escape{n} England {\color{stage-green} Football} Association chairman Stephen McClaren confirmed the {\color{stage-green} acquisition} of the 20-year-old England international on Wednesday, making the move official following a lucrative {\color{stage-green} transfer} agreement with {\color{stage-orange} Liverpool}, {\color{stage-green} {\color{stage-orange}Bayern}  Munich} and {\color{stage-orange} Chelsea}.\escape{n} {\color{stage-orange} Beckham} spent the first half of this year on loan at {\color{stage-orange} Chelsea} and did not start a single match during the {\color{stage-green} season} at the {\color{stage-green} Italian} capital's San Siro despite being named the player of the {\color{stage-green} season} of the year by {\color{stage-green} Italy}. \escape{n} However, he played 17 times last {\color{stage-green} season} and 12 times this season for {\color{stage-orange} Liverpool}, making 33 league appearances and scoring seven goals. \escape{n} He is currently third on the all-time England goalscoring list behind only England manager Alex Ferguson and new {\color{stage-orange} club} teammate Paul Gascoigne. [...]
 \\

\bottomrule
\end{tabular}
\vspace{-8pt}
    \caption{An excerpt of a 3-stage generated example by \PROG-3 on the CNN News domain.}
    \label{tab:control_1}
    \vspace{-8pt}
\end{table*}

\subsection{Human Evaluation}
In our human study, we asked three university students who are proficient English speakers to evaluate the coherence and fluency of the generated text. To better assess the coherence of the long passages of text, we evaluate at both the passage level and the finer-grained sentence level. More concretely, for {\bf passage-level coherence}, human raters assign a coherence score to each full-length text sample, on a 5-point Likert scale. For a more detailed assessment, we further evaluate {\bf sentence-level coherence}, where human raters label each sentence in the text passage with 0 or 1, indicating whether the particular sentence is coherent with the proceeding context in the passage. We then calculate the average percentage of coherent sentences in the generated text by each model. Human raters also evaluate the language quality for a {\bf fluency} score on a 5-point Likert scale. We compare our method with the systems that show highest generation quality in automatic evaluation, including \texttt{BART}, \texttt{GPT2-Small}, and \texttt{Summary}.
We evaluated 50 examples for each comparison model on the CNN domain. The Pearson correlation coefficient of human scores is 0.52, showing moderate inter-rater agreement.

Table~\ref{tab:human_eval} shows the results. All systems receive close fluency scores. Our approach obtained significantly higher coherence scores at both passage and sentence levels. In particular, over 86\% sentences in our model generations are considered as coherent with the context, improving over other models by at least 10 absolute percent.


\subsection{Ablation Study and Analysis}

\paragraph{Sample efficiency.}
We study how the progressive generation could improve the sample efficiency of large LMs fine-tuned to target domains. The intuition is that by focusing on the subsets of informative words, the early stages can more efficiently capture the domain-specific characteristics and then steer the subsequent refinement stages. Figure~\ref{fig:sample_efficiency} shows the results where we report the FBD score averaged over FBD-S/M/D. We can see our approach can make more efficient use of the training data in learning to generate high quality samples.
For example, with only 1K training examples, our method achieves comparable results with large LMs trained on 30K examples.

\paragraph{Generation with gold plans.}

To investigate the importance of dividing the generation process into stages and what the stages learn separately, we add another set of text into our comparison. It is a 2-stages model whose first stage is the ground truth (gold plan) while the second stage kept the same (a BART model), shown as \texttt{GoldPlan} in Table~\ref{tab:gold_plan}. Note that
with gold plan, our model greatly decreases the gap with human text in terms of lexical (TID) and semantic (FBD-D) quality metrics. The results highlight the importance of plans in text generation. The intermediate
plans act as an information bottleneck, and high-quality plans could lead to high-quality text generation.

\paragraph{Effect of data noising.}
We study the ablation of data noising, to check whether the noising operation really helps 
reduce stage-wise exposure bias (Sec~\ref{sec:method:train}) as we expected. 
Table~\ref{tab:noise} shows the comparison between models with and without noise in 
training. The added noise generally brings performance improvement in terms of various metrics.

\paragraph{Example generations.}

Table~\ref{tab:control_1} shows an example of text generated via three stages. We can
see our model first generates the key subject {\em beckham} and the team name {\em liverpool} in
the very first stage, then adds more fine-grained details like {\em acquisition, transfer} in
the second stage and finally expands the keywords into a full document describing Beckham's joining
a new team.

\section{Conclusion}
\label{sec:conclusion}

We have proposed a new approach for domain-specific generation of long text passages in a progressive manner.
Our method is simple and efficient by fine-tuning large-scale off-the-shelf language models. We conduct extensive experiments using a variety of metrics and human studies.
We demonstrate that our method outperforms a wide range of large pretrained LMs with single-stage generation or prior planning-then-generation strategies, in terms of quality and coherence of the produced samples.
The multi-stage generation also opens up new opportunities to enhance controllability of text generation, which we would love to explore in the future. 

\bibliography{anthology,custom}

\begin{thebibliography}{48}
\expandafter\ifx\csname natexlab\endcsname\relax\def\natexlab#1{#1}\fi

\bibitem[{Bosselut et~al.(2018)Bosselut, Celikyilmaz, He, Gao, Huang, and
  Choi}]{bosselut2018discourse}
Antoine Bosselut, Asli Celikyilmaz, Xiaodong He, Jianfeng Gao, Po-Sen Huang,
  and Yejin Choi. 2018.
\newblock Discourse-aware neural rewards for coherent text generation.
\newblock In \emph{NAACL}, pages 173--184.

\bibitem[{Brown et~al.(2020)Brown, Mann, Ryder, Subbiah, Kaplan, Dhariwal,
  Neelakantan, Shyam, Sastry, Askell et~al.}]{brown2020language}
Tom~B Brown, Benjamin Mann, Nick Ryder, Melanie Subbiah, Jared Kaplan, Prafulla
  Dhariwal, Arvind Neelakantan, Pranav Shyam, Girish Sastry, Amanda Askell,
  et~al. 2020.
\newblock \href
  {https://proceedings.neurips.cc/paper/2020/file/1457c0d6bfcb4967418bfb8ac142f64a-Paper.pdf}
  {Language models are few-shot learners}.
\newblock In \emph{NeurIPS}, volume~33, pages 1877--1901.

\bibitem[{Chan et~al.(2019)Chan, Kitaev, Guu, Stern, and
  Uszkoreit}]{chan2019kermit}
William Chan, Nikita Kitaev, Kelvin Guu, Mitchell Stern, and Jakob Uszkoreit.
  2019.
\newblock {KERMIT}: Generative insertion-based modeling for sequences.
\newblock \emph{arXiv preprint arXiv:1906.01604}.

\bibitem[{Choromanski et~al.(2021)Choromanski, Likhosherstov, Dohan, Song,
  Gane, Sarlos, Hawkins, Davis, Mohiuddin, Kaiser
  et~al.}]{choromanski2020rethinking}
Krzysztof Choromanski, Valerii Likhosherstov, David Dohan, Xingyou Song,
  Andreea Gane, Tamas Sarlos, Peter Hawkins, Jared Davis, Afroz Mohiuddin,
  Lukasz Kaiser, et~al. 2021.
\newblock Rethinking attention with performers.
\newblock \emph{ICLR}.

\bibitem[{Dai et~al.(2019)Dai, Yang, Yang, Carbonell, Le, and
  Salakhutdinov}]{dai2019transformer}
Zihang Dai, Zhilin Yang, Yiming Yang, Jaime Carbonell, Quoc Le, and Ruslan
  Salakhutdinov. 2019.
\newblock \href {https://doi.org/10.18653/v1/P19-1285} {Transformer-{XL}:
  Attentive language models beyond a fixed-length context}.
\newblock In \emph{ACL}, pages 2978--2988.

\bibitem[{Devlin et~al.(2019)Devlin, Chang, Lee, and
  Toutanova}]{devlin2018bert}
Jacob Devlin, Ming-Wei Chang, Kenton Lee, and Kristina Toutanova. 2019.
\newblock \href {https://doi.org/10.18653/v1/N19-1423} {{BERT}: Pre-training of
  deep bidirectional transformers for language understanding}.
\newblock In \emph{NAACL}, pages 4171--4186.

\bibitem[{Fan et~al.(2018)Fan, Lewis, and Dauphin}]{fan2018hierarchical}
Angela Fan, Mike Lewis, and Yann Dauphin. 2018.
\newblock \href {https://doi.org/10.18653/v1/P18-1082} {Hierarchical neural
  story generation}.
\newblock In \emph{ACL}, pages 889--898.

\bibitem[{Fan et~al.(2019)Fan, Lewis, and Dauphin}]{fan2019strategies}
Angela Fan, Mike Lewis, and Yann Dauphin. 2019.
\newblock Strategies for structuring story generation.
\newblock In \emph{ACL}.

\bibitem[{Ford et~al.(2018)Ford, Duckworth, Norouzi, and
  Dahl}]{ford2018importance}
Nicolas Ford, Daniel Duckworth, Mohammad Norouzi, and George~E Dahl. 2018.
\newblock The importance of generation order in language modeling.
\newblock In \emph{EMNLP}.

\bibitem[{Gardian()}]{gpt3guardian}
Gardian.
\newblock \href
  {https://www.theguardian.com/commentisfree/2020/sep/08/robot-wrote-this-article-gpt-3}
  {A robot wrote this entire article. are you scared yet, human?}

\bibitem[{Gu et~al.(2019)Gu, Liu, and Cho}]{gu2019insertion}
Jiatao Gu, Qi~Liu, and Kyunghyun Cho. 2019.
\newblock Insertion-based decoding with automatically inferred generation
  order.
\newblock \emph{TACL}, 7:661--676.

\bibitem[{Hermann et~al.(2015)Hermann, Kocisky, Grefenstette, Espeholt, Kay,
  Suleyman, and Blunsom}]{hermann2015teaching}
Karl~Moritz Hermann, Tomas Kocisky, Edward Grefenstette, Lasse Espeholt, Will
  Kay, Mustafa Suleyman, and Phil Blunsom. 2015.
\newblock Teaching machines to read and comprehend.
\newblock In \emph{NeurIPS}, pages 1693--1701.

\bibitem[{Holtzman et~al.(2020)Holtzman, Buys, Forbes, and
  Choi}]{holtzman2019curious}
Ari Holtzman, Jan Buys, Maxwell Forbes, and Yejin Choi. 2020.
\newblock The curious case of neural text degeneration.
\newblock In \emph{ICLR}.

\bibitem[{Hua and Wang(2019)}]{hua2019sentence}
Xinyu Hua and Lu~Wang. 2019.
\newblock Sentence-level content planning and style specification for neural
  text generation.
\newblock In \emph{EMNLP}.

\bibitem[{Hua and Wang(2020)}]{hua2020pair}
Xinyu Hua and Lu~Wang. 2020.
\newblock {PAIR}: Planning and iterative refinement in pre-trained transformers
  for long text generation.
\newblock In \emph{EMNLP}, pages 781--793.

\bibitem[{Kasai et~al.(2020)Kasai, Cross, Ghazvininejad, and
  Gu}]{kasai2020disCo}
Jungo Kasai, James Cross, Marjan Ghazvininejad, and Jiatao Gu. 2020.
\newblock \href {https://arxiv.org/abs/2001.05136} {Non-autoregressive machine
  translation with disentangled context transformer}.
\newblock In \emph{ICML}.

\bibitem[{Lee et~al.(2018)Lee, Mansimov, and Cho}]{lee2018deterministic}
Jason Lee, Elman Mansimov, and Kyunghyun Cho. 2018.
\newblock \href {https://doi.org/10.18653/v1/D18-1149} {Deterministic
  non-autoregressive neural sequence modeling by iterative refinement}.
\newblock In \emph{EMNLP}, pages 1173--1182.

\bibitem[{Lewis et~al.(2020)Lewis, Liu, Goyal, Ghazvininejad, Mohamed, Levy,
  Stoyanov, and Zettlemoyer}]{lewis2019bart}
Mike Lewis, Yinhan Liu, Naman Goyal, Marjan Ghazvininejad, Abdelrahman Mohamed,
  Omer Levy, Veselin Stoyanov, and Luke Zettlemoyer. 2020.
\newblock \href {https://doi.org/10.18653/v1/2020.acl-main.703} {{BART}:
  Denoising sequence-to-sequence pre-training for natural language generation,
  translation, and comprehension}.
\newblock In \emph{ACL}, pages 7871--7880.

\bibitem[{Liu et~al.(2018)Liu, Saleh, Pot, Goodrich, Sepassi, Kaiser, and
  Shazeer}]{liu2018generating}
Peter~J Liu, Mohammad Saleh, Etienne Pot, Ben Goodrich, Ryan Sepassi, Lukasz
  Kaiser, and Noam Shazeer. 2018.
\newblock Generating wikipedia by summarizing long sequences.
\newblock In \emph{ICLR}.

\bibitem[{Mansimov et~al.(2019)Mansimov, Wang, Welleck, and
  Cho}]{mansimov2019generalized}
Elman Mansimov, Alex Wang, Sean Welleck, and Kyunghyun Cho. 2019.
\newblock A generalized framework of sequence generation with application to
  undirected sequence models.
\newblock \emph{arXiv preprint arXiv:1905.12790}.

\bibitem[{MarketMuse()}]{gpt3exposed}
MarketMuse.
\newblock \href {https://blog.marketmuse.com/gpt-3-exposed/} {Gpt-3 exposed:
  Behind the smoke and mirrors}.

\bibitem[{Montahaei et~al.(2019)Montahaei, Alihosseini, and
  Baghshah}]{montahaei2019jointly}
Ehsan Montahaei, Danial Alihosseini, and Mahdieh~Soleymani Baghshah. 2019.
\newblock Jointly measuring diversity and quality in text generation models.
\newblock \emph{NAACL Workshop}.

\bibitem[{Moryossef et~al.(2019)Moryossef, Goldberg, and
  Dagan}]{moryossef2019step}
Amit Moryossef, Yoav Goldberg, and Ido Dagan. 2019.
\newblock Step-by-step: Separating planning from realization in neural
  data-to-text generation.
\newblock In \emph{NAACL}.

\bibitem[{Novak et~al.(2016)Novak, Auli, and Grangier}]{novak2016iterative}
Roman Novak, Michael Auli, and David Grangier. 2016.
\newblock Iterative refinement for machine translation.
\newblock \emph{arXiv preprint arXiv:1610.06602}.

\bibitem[{Puduppully et~al.(2019)Puduppully, Dong, and
  Lapata}]{puduppully2019data}
Ratish Puduppully, Li~Dong, and Mirella Lapata. 2019.
\newblock Data-to-text generation with content selection and planning.
\newblock In \emph{AAAI}, volume~33, pages 6908--6915.

\bibitem[{Qin et~al.(2020)Qin, Shwartz, West, Bhagavatula, Hwang, Le~Bras,
  Bosselut, and Choi}]{qin-etal-2020-back}
Lianhui Qin, Vered Shwartz, Peter West, Chandra Bhagavatula, Jena~D. Hwang,
  Ronan Le~Bras, Antoine Bosselut, and Yejin Choi. 2020.
\newblock \href {https://doi.org/10.18653/v1/2020.emnlp-main.58} {Back to the
  future: Unsupervised backprop-based decoding for counterfactual and abductive
  commonsense reasoning}.
\newblock In \emph{EMNLP}, pages 794--805. Association for Computational
  Linguistics.

\bibitem[{Radford et~al.(2019)Radford, Wu, Child, Luan, Amodei, and
  Sutskever}]{radford2019language}
Alec Radford, Jeffrey Wu, Rewon Child, David Luan, Dario Amodei, and Ilya
  Sutskever. 2019.
\newblock Language models are unsupervised multitask learners.
\newblock \emph{OpenAI Blog}, 1(8):9.

\bibitem[{Ranzato et~al.(2016)Ranzato, Chopra, Auli, and
  Zaremba}]{ranzato2015sequence}
Marc'Aurelio Ranzato, Sumit Chopra, Michael Auli, and Wojciech Zaremba. 2016.
\newblock Sequence level training with recurrent neural networks.
\newblock \emph{ICLR}.

\bibitem[{Rashkin et~al.(2020)Rashkin, Celikyilmaz, Choi, and
  Gao}]{rashkin-etal-2020-plotmachines}
Hannah Rashkin, Asli Celikyilmaz, Yejin Choi, and Jianfeng Gao. 2020.
\newblock \href {https://www.aclweb.org/anthology/2020.emnlp-main.349}
  {{P}lot{M}achines: Outline-conditioned generation with dynamic plot state
  tracking}.
\newblock In \emph{EMNLP}, pages 4274--4295.

\bibitem[{Reiter and Dale(1997)}]{reiter1997building}
Ehud Reiter and Robert Dale. 1997.
\newblock Building applied natural language generation systems.
\newblock \emph{Natural Language Engineering}, 3(1):57--87.

\bibitem[{Rose et~al.(2010)Rose, Engel, Cramer, and Cowley}]{rose2010automatic}
Stuart Rose, Dave Engel, Nick Cramer, and Wendy Cowley. 2010.
\newblock Automatic keyword extraction from individual documents.
\newblock \emph{Text mining: applications and theory}, 1:1--20.

\bibitem[{Salton and McGill(1986)}]{salton1986introduction}
Gerard Salton and Michael~J McGill. 1986.
\newblock Introduction to modern information retrieval.

\bibitem[{See et~al.(2019)See, Pappu, Saxena, Yerukola, and
  Manning}]{see2019massively}
Abigail See, Aneesh Pappu, Rohun Saxena, Akhila Yerukola, and Christopher~D.
  Manning. 2019.
\newblock \href {https://doi.org/10.18653/v1/K19-1079} {Do massively pretrained
  language models make better storytellers?}
\newblock In \emph{CoNLL}, pages 843--861.

\bibitem[{Shen et~al.(2019)Shen, Celikyilmaz, Zhang, Chen, Wang, Gao, and
  Carin}]{shen2019towards}
Dinghan Shen, Asli Celikyilmaz, Yizhe Zhang, Liqun Chen, Xin Wang, Jianfeng
  Gao, and Lawrence Carin. 2019.
\newblock Towards generating long and coherent text with multi-level latent
  variable models.
\newblock In \emph{ACL}, pages 2079--2089.

\bibitem[{Shen et~al.(2020)Shen, Quach, Barzilay, and Jaakkola}]{shen2020blank}
Tianxiao Shen, Victor Quach, Regina Barzilay, and Tommi Jaakkola. 2020.
\newblock \href {https://doi.org/10.18653/v1/2020.emnlp-main.420} {Blank
  language models}.
\newblock In \emph{EMNLP}, pages 5186--5198.

\bibitem[{Shi et~al.(2018)Shi, Chen, Qiu, and Huang}]{shi2018toward}
Zhan Shi, Xinchi Chen, Xipeng Qiu, and Xuanjing Huang. 2018.
\newblock Toward diverse text generation with inverse reinforcement learning.
\newblock \emph{IJCAI}.

\bibitem[{Stern et~al.(2019)Stern, Chan, Kiros, and
  Uszkoreit}]{stern2019insertion}
Mitchell Stern, William Chan, Jamie Kiros, and Jakob Uszkoreit. 2019.
\newblock \href {http://proceedings.mlr.press/v97/stern19a.html} {Insertion
  transformer: Flexible sequence generation via insertion operations}.
\newblock In \emph{ICML}, volume~97, pages 5976--5985.

\bibitem[{Tan et~al.(2019)Tan, Hu, Yang, Salakhutdinov, and
  Xing}]{tan2018connecting}
Bowen Tan, Zhiting Hu, Zichao Yang, Ruslan Salakhutdinov, and Eric~P Xing.
  2019.
\newblock Connecting the dots between mle and rl for sequence generation.
\newblock \emph{ICLR Workshop}.

\bibitem[{Vaswani et~al.(2017)Vaswani, Shazeer, Parmar, Uszkoreit, Jones,
  Gomez, Kaiser, and Polosukhin}]{vaswani2017attention}
Ashish Vaswani, Noam Shazeer, Niki Parmar, Jakob Uszkoreit, Llion Jones,
  Aidan~N Gomez, {\L}ukasz Kaiser, and Illia Polosukhin. 2017.
\newblock Attention is all you need.
\newblock In \emph{NeurIPS}, pages 5998--6008.

\bibitem[{Wang et~al.(2020)Wang, Li, Khabsa, Fang, and Ma}]{wang2020linformer}
Sinong Wang, Belinda Li, Madian Khabsa, Han Fang, and Hao Ma. 2020.
\newblock Linformer: Self-attention with linear complexity.
\newblock \emph{arXiv preprint arXiv:2006.04768}.

\bibitem[{Welleck et~al.(2019)Welleck, Brantley, Daum{\'e}~III, and
  Cho}]{welleck2019non}
Sean Welleck, Kiant{\'e} Brantley, Hal Daum{\'e}~III, and Kyunghyun Cho. 2019.
\newblock Non-monotonic sequential text generation.
\newblock In \emph{ICML}.

\bibitem[{Xie et~al.(2017)Xie, Wang, Li, L{\'e}vy, Nie, Jurafsky, and
  Ng}]{xie2017data}
Ziang Xie, Sida~I Wang, Jiwei Li, Daniel L{\'e}vy, Aiming Nie, Dan Jurafsky,
  and Andrew~Y Ng. 2017.
\newblock Data noising as smoothing in neural network language models.
\newblock In \emph{ICLR}.

\bibitem[{Xu et~al.(2020)Xu, Patwary, Shoeybi, Puri, Fung, Anandkumar, and
  Catanzaro}]{xu2020megatron}
Peng Xu, Mostofa Patwary, Mohammad Shoeybi, Raul Puri, Pascale Fung, Anima
  Anandkumar, and Bryan Catanzaro. 2020.
\newblock \href {https://doi.org/10.18653/v1/2020.emnlp-main.226}
  {{MEGATRON}-{CNTRL}: Controllable story generation with external knowledge
  using large-scale language models}.
\newblock In \emph{EMNLP}, pages 2831--2845.

\bibitem[{Yao et~al.(2019)Yao, Peng, Weischedel, Knight, Zhao, and
  Yan}]{yao2019plan}
Lili Yao, Nanyun Peng, Ralph Weischedel, Kevin Knight, Dongyan Zhao, and Rui
  Yan. 2019.
\newblock Plan-and-write: Towards better automatic storytelling.
\newblock In \emph{AAAI}, volume~33, pages 7378--7385.

\bibitem[{Yao et~al.(2017)Yao, Zhang, Feng, Zhao, and Yan}]{yao2017towards}
Lili Yao, Yaoyuan Zhang, Yansong Feng, Dongyan Zhao, and Rui Yan. 2017.
\newblock Towards implicit content-introducing for generative short-text
  conversation systems.
\newblock In \emph{EMNLP}, pages 2190--2199.

\bibitem[{Zhang et~al.(2020)Zhang, Wang, Li, Gan, Brockett, and
  Dolan}]{zhang2020pointer}
Yizhe Zhang, Guoyin Wang, Chunyuan Li, Zhe Gan, Chris Brockett, and Bill Dolan.
  2020.
\newblock \href {https://www.aclweb.org/anthology/2020.emnlp-main.698}
  {{POINTER}: Constrained progressive text generation via insertion-based
  generative pre-training}.
\newblock In \emph{EMNLP}, pages 8649--8670.

\bibitem[{Zhao et~al.(2020)Zhao, Xu, Lin, Zhang, Yang, and Sun}]{zhao2020graph}
Liang Zhao, Jingjing Xu, Junyang Lin, Yichang Zhang, Hongxia Yang, and Xu~Sun.
  2020.
\newblock Graph-based multi-hop reasoning for long text generation.
\newblock \emph{arXiv preprint arXiv:2009.13282}.

\bibitem[{Zhu et~al.(2019)Zhu, Hu, and Xing}]{zhu2019text}
Wanrong Zhu, Zhiting Hu, and Eric Xing. 2019.
\newblock Text infilling.
\newblock \emph{arXiv preprint arXiv:1901.00158}.

\end{thebibliography}
\bibliographystyle{acl_natbib}

\onecolumn

\section*{Appendix: Complete Results}

We include complete result numbers of experiments here.

\begin{table}[h]
\small
    \centering
\begin{tabular}{@{}lllllllllllll@{}}
\toprule
       & GPT2-S & GPT2-L & BART  & Summ. & RAKE  & SRL   & SRL-N & SRL-C & \PROG-2 & \PROG-3 & \PROG-4 & Dev   \\ \midrule
B-BL2  & 72.84  & 71.89  & 71.51 & 73.28 & 69.78 & 70.25 & 74.50 & 74.71 & 72.25    & 74.10    & 74.57    & 75.82 \\
B-BL3  & 48.53  & 47.48  & 47.55 & 49.26 & 45.39 & 46.54 & 51.19 & 51.40 & 48.44    & 50.38    & 51.06    & 52.08 \\
B-BL4  & 28.64  & 28.55  & 28.11 & 29.31 & 26.09 & 27.25 & 31.04 & 31.06 & 28.88    & 30.32    & 30.96    & 32.29 \\
B-BL5  & 15.87  & 15.62  & 15.57 & 16.35 & 14.01 & 14.88 & 17.58 & 17.41 & 16.08    & 17.09    & 17.53    & 19.35 \\ \midrule
HA-BL2 & 73.61  & 71.97  & 74.56 & 74.59 & 71.63 & 67.47 & 74.51 & 75.11 & 74.64    & 75.17    & 75.86    & 75.72 \\
HA-BL3 & 49.26  & 47.83  & 50.27 & 50.32 & 47.34 & 44.51 & 50.87 & 51.18 & 50.64    & 51.07    & 51.88    & 52.01 \\
HA-BL4 & 29.21  & 28.26  & 30.03 & 29.88 & 27.51 & 25.84 & 30.45 & 30.49 & 30.45    & 30.64    & 31.32    & 32.28 \\
HA-BL5 & 16.22  & 15.77  & 16.77 & 16.52 & 14.84 & 13.91 & 16.94 & 16.87 & 17.09    & 17.18    & 17.63    & 19.40 \\ \midrule
MSJ-2  & 49.24  & 46.94  & 49.85 & 46.97 & 44.19 & 43.85 & 49.39 & 44.37 & 49.46    & 50.16    & 51.00    & 54.51 \\
MSJ-3  & 28.79  & 27.29  & 29.43 & 27.99 & 26.01 & 25.90 & 29.58 & 26.92 & 29.54    & 30.04    & 30.56    & 32.54 \\
MSJ-4  & 15.73  & 14.85  & 16.24 & 15.48 & 14.12 & 14.15 & 16.33 & 14.99 & 16.50    & 16.68    & 16.96    & 18.60 \\
MSJ-5  & 8.38   & 7.91   & 8.72  & 8.25  & 7.36  & 7.43  & 8.68  & 8.02  & 8.90     & 8.95     & 9.10     & 10.87 \\ \midrule
TID    & 8.7    & 9.2    & 6.8   & 4.5   & 7.8   & 16.1  & 5.2   & 5.2   & 6.2      & 5.4      & 4.0      & 2.6   \\ \midrule
FBD-S  & 16.21  & 18.50  & 7.76  & 2.93  & 4.17  & 14.26 & 11.42 & 4.66  & 3.26     & 3.16     & 2.64     & 5.98  \\
FBD-M  & 24.92  & 29.61  & 22.49 & 15.00 & 25.92 & 37.24 & 22.63 & 20.28 & 19.05    & 18.84    & 17.38    & 12.26 \\
FBD-D  & 43.07  & 44.15  & 44.86 & 33.08 & 54.12 & 64.83 & 43.26 & 44.34 & 39.94    & 38.30    & 36.49    & 25.63 \\ \bottomrule
\end{tabular}
    \caption{Complete results on the CNN News domain.}
    \label{tab:my_label}
\end{table}


\begin{table}[h]
\small
    \centering
\begin{tabular}{@{}lllllllllllll@{}}
\toprule
       & GPT2-S & GPT2-L & BART  & Summ. & RAKE   & SRL   & SRL-N & SRL-C & ProGet-2 & ProGet-3 & ProGet-4 & Dev   \\ \midrule
B-BL2  & 78.38  & 77.43  & 76.96 & 77.19 & 76.97  & 77.98 & 77.90 & 77.62 & 78.64    & 78.73    & 78.41    & 79.20 \\
B-BL3  & 55.51  & 54.18  & 54.45 & 54.45 & 53.86  & 55.67 & 55.49 & 55.09 & 56.44    & 56.50    & 56.25    & 56.02 \\
B-BL4  & 33.41  & 32.20  & 33.02 & 32.88 & 31.95  & 33.83 & 33.75 & 33.36 & 34.46    & 34.62    & 34.52    & 34.08 \\
B-BL5  & 17.59  & 16.79  & 17.55 & 17.53 & 16.47  & 17.93 & 17.98 & 17.63 & 18.32    & 18.49    & 18.57    & 18.40 \\ \midrule
HA-BL2 & 78.19  & 76.96  & 79.99 & 79.30 & 77.19  & 79.24 & 77.73 & 77.46 & 80.57    & 80.72    & 80.50    & 79.51 \\
HA-BL3 & 55.39  & 54.33  & 57.86 & 56.83 & 54.71  & 57.00 & 55.71 & 55.14 & 58.11    & 58.38    & 58.35    & 56.39 \\
HA-BL4 & 33.32  & 32.52  & 35.63 & 34.63 & 32.70  & 34.63 & 33.93 & 33.36 & 35.43    & 35.84    & 35.96    & 34.36 \\
HA-BL5 & 17.46  & 16.94  & 19.16 & 18.47 & 16.86  & 18.26 & 18.03 & 17.60 & 18.72    & 19.14    & 19.30    & 18.55 \\ \midrule
MSJ-2  & 55.27  & 54.21  & 55.89 & 52.63 & 51..88 & 47.51 & 45.39 & 43.36 & 55.14    & 56.51    & 56.18    & 60.07 \\
MSJ-3  & 34.48  & 33.70  & 35.46 & 33.46 & 32.59  & 30.88 & 29.51 & 28.22 & 34.81    & 35.80    & 35.74    & 37.42 \\
MSJ-4  & 19.32  & 18.83  & 20.27 & 19.17 & 18.33  & 17.87 & 17.11 & 16.39 & 19.63    & 20.29    & 20.39    & 21.22 \\
MSJ-5  & 10.16  & 9.90   & 10.73 & 10.27 & 9.57   & 9.54  & 9.21  & 8.82  & 10.16    & 10.60    & 10.76    & 11.34 \\ \midrule
TID    & 4.6    & 8.3    & 5.1   & 4.5   & 5.8    & 5.5   & 5.3   & 7.0   & 5.1      & 5.0      & 4.8      & 3.4   \\ \midrule
FBD-S  & 3.49   & 3.43   & 5.34  & 5.06  & 8.28   & 6.03  & 7.49  & 8.63  & 3.72     & 3.90     & 3.81     & 1.96  \\
FBD-M  & 19.30  & 19.41  & 21.75 & 18.11 & 22.97  & 21.85 & 23.15 & 25.01 & 19.36    & 19.04    & 18.62    & 12.23 \\
FBD-D  & 40.18  & 41.22  & 43.97 & 33.90 & 44.32  & 43.63 & 45.87 & 48.92 & 39.82    & 39.05    & 38.68    & 28.82 \\ \bottomrule
\end{tabular}
    \caption{Complete results on the story domain.}
    \label{tab:my_label}
\end{table}





\end{document}